\documentclass[10pt,journal,compsoc]{IEEEtran}

\usepackage{amsmath}
\usepackage{diagbox}
\usepackage{graphics}
\usepackage{epsfig}
\usepackage{threeparttable}
\usepackage{xspace}
\usepackage{siunitx}
\usepackage{graphicx}
\usepackage{subfig}
\usepackage{amssymb}
\usepackage{threeparttable}
\usepackage{color}
\usepackage[normalem]{ulem}
\usepackage{multirow}
\usepackage{float}
\usepackage{amsfonts}
\usepackage{bm}
\usepackage{array}
\usepackage[table]{xcolor}
\usepackage{colortbl}
\usepackage{diagbox}
\usepackage{rotating}
\usepackage{booktabs}
\usepackage{overpic}
\usepackage{enumitem}
\usepackage{colortbl}
\usepackage{algorithm}
\usepackage{algorithmic}
\usepackage{cite}
\usepackage{dblfloatfix}
\usepackage[export]{adjustbox}
\usepackage{pifont}
\usepackage[pagebackref=false,breaklinks=true,colorlinks,bookmarks=false]{hyperref}
\usepackage{makecell}

\begin{document}

\title{FMLGS: Fast Multilevel Language Embedded Gaussians for Part-level Interactive Agents}

\author{Xin Tan, Yuzhou Ji, He Zhu, and Yuan Xie

\thanks{Manuscript received xx xx 2025; revised xx xx 2025; accepted xx xx 2025.
This work is supported by the National Natural Science Foundation of China No.62302296, U23A20343, 62176092, 62106075, 62302167, Shanghai Sailing Program (23YF1410500), Chenguang Program of Shanghai Education Development Foundation and Shanghai Municipal Education Commission (23CGA34), Natural Science Foundation of Chongqing, China (CSTB2023NSCQ-JQX0007, CSTB2023NSCQ-MSX0137), and Development Project of Ministry of Industry and Information Technology (ZTZB.23-990-016). (Xin Tan and Yuzhou Ji contribute equally) (Corresponding author: Yuan Xie.)}

\thanks{Xin Tan, Yuzhou Ji, He Zhu, and Yuan Xie are with the School of Computer Science and Technology, East China Normal University, Shanghai 200062, China (e-mails: xtan@cs.ecnu.edu.cn; 10215102492@stu.ecnu.edu.cn; 10215102469@stu.ecnu.edu.cn; xieyuan8589@foxmail.com).}
\thanks{Yuan Xie is also with the Shanghai Innovation Institute, Shanghai 200062, China.}
}


\markboth{Journal of \LaTeX\ Class Files,~Vol.~14, No.~8, August~2021}%
{Shell \MakeLowercase{\textit{et al.}}: A Sample Article Using IEEEtran.cls for IEEE Journals}


\maketitle

\begin{abstract}
The semantically interactive radiance field has long been a promising backbone for 3D real-world applications, such as embodied AI to achieve scene understanding and manipulation. 
However, multigranularity interaction remains a challenging task due to the ambiguity of language and degraded quality when it comes to queries upon object components. 
In this work, we present FMLGS, an approach that supports part-level open-vocabulary query within 3D Gaussian Splatting (3DGS). 
We propose an efficient pipeline for building and querying consistent object- and part-level semantics based on Segment Anything Model 2 (SAM2). 
We designed a semantic deviation strategy to solve the problem of language ambiguity among object parts, which interpolates the semantic features of fine-grained targets for enriched information. 
Once trained, we can query both objects and their describable parts using natural language. 
Comparisons with other state-of-the-art methods prove that our method can not only better locate specified part-level targets, but also achieve first-place performance concerning both \textbf{speed} and \textbf{accuracy}, where FMLGS is 98 $\times$ faster than LERF, 4 $\times$ faster than LangSplat and 2.5 $\times$ faster than LEGaussians. 
Meanwhile, we further integrate FMLGS as a virtual agent that can interactively navigate through 3D scenes, locate targets, and respond to user demands through a chat interface, which demonstrates the potential of our work to be further expanded and applied in the future.
\end{abstract}


\section{Introduction}
\IEEEPARstart{A}{s} a specially reconstructed 3D scene, language-embedded radiance fields can be semantically interactive under continuous new visual angles. Therefore, it serves as a promising component within fields such as embodied AI and augmented reality. 
Given a set of posed images, a language-embedded radiance field learns an accurate and efficient 3D representation of semantics, enabling scene understanding and manipulation by providing consistent semantics across any views. 
Although much progress has been made in object-level language-embedded radiance fields, it still
remains a vital challenge to achieve high-quality part-level semantics. 

\begin{figure}[t]
   \centering
   \includegraphics[width = 1.0\columnwidth]{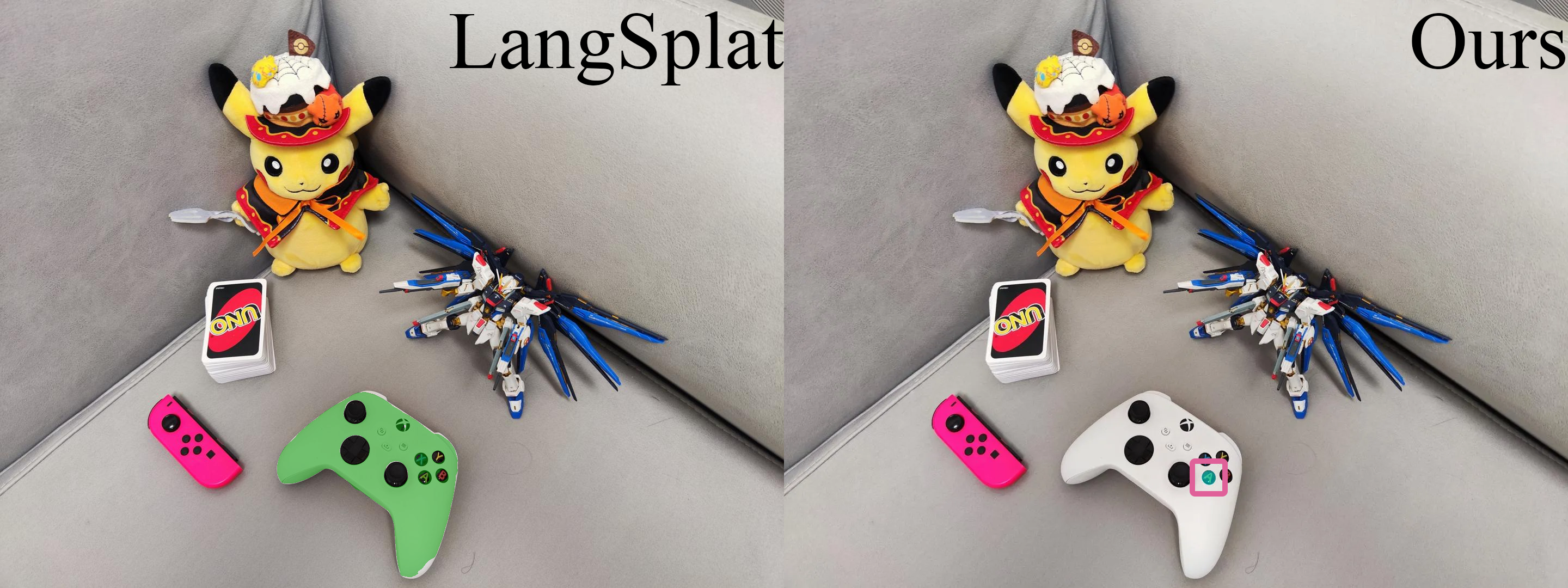}
   \caption{Results of querying ``A Button of Xbox Wireless Controller''. While LangSplat \cite{qin2023langsplat} fails in detailed part-level localization, our method provides an accurate outcome.}
   \label{fig:intro}
\end{figure}

The existing works mainly lie on pixel-based and mask-based methods. For pixel-based methods, they are good at detecting targets of either whole objects or small parts of interest, but cannot provide clear target outline, limiting further applications concerning demanding 3D manipulation tasks. 
For example, LERF \cite{Kerr_2023_ICCV} firstly enables pixel-aligned queries of the distilled 3D CLIP \cite{radford2021learning} embeddings, supporting long-tail open-vocabulary queries within NeRF \cite{mildenhall2021nerf}. 
Although LERF's heat map shows astonishing localization ability upon object parts such as ``bear nose'' and ``engine'', the results are too fuzzy to be used for downstream applications. 
Based on LERF's core concept, LEGaussians \cite{shi2023language} proposes a feature lifting strategy within 3DGS \cite{kerbl3Dgaussians}, achieving faster queries and smoother relevancy maps, but the results are still unstable and inconsistent. 

In contrast, the mask-based methods perform well in generating accurate query masks while suffering from limitedly recallable targets. 
For example, Gaussian Grouping \cite{ye2023gaussian} introduces identity encoding and 3D spatial consistency regularization, consistently lifting 2D Segment Anything Model (SAM) \cite{Kirillov_2023_ICCV} masks into 3D scenes and provides high-quality object-level results. 
In FastLGS \cite{ji2025fastlgs}, we also propose cross-view grid mapping to achieve accurate results and fast queries. 
While these methods are based solely on \textit{everything} mode SAM masks to ensure consistent object identity, they lose the ability to query any object parts. 
LangSplat \cite{qin2023langsplat}, on the other hand, learns hierarchical semantics by sorting and filtering SAM masks into \textit{whole}, \textit{part} and \textit{subpart} groups for lifting multi-scale language embeddings. 
This technique enables LangSplat to further provide query ability upon certain components such as ``yellow'' parts of a Pikachu toy while still having high accuracy in object-level results. 
However, LangSplat only processes masks generated from the same 32 $\times$ 32 point prompts, supporting extremely limited partial queries. 
Meanwhile, it uses the raw CLIP embeddings which have no subordination information and also degrade after 
dimensionality reduction, resulting in failure concerning detailed part-level queries (see Figure \ref{fig:intro}). 

The above discussion shows the dilemma of balancing query accuracy and 
part-level localization ability when building a language-embedded radiance field.

To get out of such a predicament, in this paper, we present the FMLGS, which provides both accurate query results and strong part-level localization ability that supports detailed description. 
Given a set of posed images, we obtain hierarchical semantics by processing in a subordination-compliant order. FMLGS first extracts all SAM masks in a single frame and filters redundantly overlapped ones to get object-level masks. Then for each object, we divide for its image tile and extract again to get masks of corresponding object parts. 
After sending them through CLIP and acquire object- and part-level embeddings separately, we designed a semantic deviation strategy to enrich object parts with subordination information. 
Then, we use SAM2 \cite{ravi2024sam2} for each object and its parts to have a consistent identity across views. 
While directly training a CLIP feature field in high dimensionality could be time-consuming, we map the features to lower 3D space based on identity, and train object- and part-level features in parallel. 
At the inference stage, open-vocabulary queries will go through both levels and generate pixel-aligned results based on relevancy scores. 

During experiments, we found FMLGS not only achieves first-place performance concerning both speed and accuracy compared with other state-of-the-art 3D segmentation and semantic field methods, but also shows the best localization ability upon specified part-level targets. 
Meanwhile, we further integrate FMLGS as a virtual agent that can interactively navigate through 3D scenes, locate targets, and respond to user demands through a chat interface, which demonstrates the potential of our work to be further expanded and applied in the future. 

In summary, the principal contributions of this work include:
\begin{itemize}
\item To our best knowledge, this work is the first to provide accurate part-level open-vocabulary localization ability among language-embedded radiance fields. 
\item We propose the multilevel feature mapping strategy with semantic deviation, which not only provides feature consistency and training efficiency, but also solves the language ambiguity issue that keeps hindering the correct target localization using natural part-level description through CLIP relevancy. 
\item We provide an effective 3D semantic basis along with an example of it being integrated into AI agents, proving promising future applications. 
\end{itemize}

\section{Related Work}
\subsection{NeRF and 3DGS}
Neural Radiance Fields (NeRF) and 3D Gaussian Splatting (3DGS) have revolutionized 3D scene modeling and rendering. After first introduction, NeRF \cite{mildenhall2021nerf} has advanced largely with further innovations from mip-nerf \cite{Barron_2021_ICCV} to NeRF-MAE \cite{irshad2024nerfmae}. 3D Gaussian Splatting, detailed by \cite{kerbl3Dgaussians} and expanded upon in work \cite{luiten2023dynamic} on dynamic 3D Gaussians, optimizes the rendering of point clouds with Gaussian kernels for real-time applications. Later contributions \cite{liu2024citygaussian,zhu2023FSGS,chen2024mvsplat,Tian2025drive} underscore the ongoing enhancements and versatility of 3DGS in handling increasingly complex rendering tasks. These works serve as a firm basis for constructing 3D language-embedded radiance fields from 2D semantics.

\subsection{2D and 3D Segmentation}
The field of 2D image segmentation has undergone remarkable progress by adopting the Transformer architecture, notably through SEgmentation TRansformer (SETR) \cite{Zheng_2021_CVPR}, alongside studies \cite{NEURIPS2021_950a4152,NEURIPS2021_64f1f27b,Cheng_2022_CVPR,sun2024uni}. 
Although works like CGRSeg \cite{ni2024context} provides excellent accuracy and efficiency, 
innovations such as SAM \cite{Kirillov_2023_ICCV} and SEEM \cite{NEURIPS2023_3ef61f7e} utilize various kinds of prompt for segmentation, inspired many mask-based researches. Meanwhile, image captioning has also become a promising method of retrieving image semantics \cite{NEURIPS2023_804b5e30}.

Advancements in 3D segmentation have paralleled those in 2D, 
with all kinds of innovations including Cylinder3D \cite{zhou2020cylinder3d} for LiDAR semantic segmentation in driving scenes, and 3D semantic segmentation within point clouds \cite{tan2023positive,sun2024image}. Other developments in radiance fields have also refined the precision and applicability of 3D segmentation techniques \cite{Goel_2023_CVPR,tang2023scene,gu2024egolifter}. 
In particular, methods based on SAM masks \cite{cen2023segment,garfield2024} achieved part-level segmentation based on user point prompts, but open-vocabulary text-guided 3D segmentation remains a challenge due to the lack of compatible semantic scene construction. 

\subsection{Language-embedded Radiance Fields}
Distilling language features into radiance fields like NeRF and 3DGS has been thoroughly explored. Zhi et al. \cite{Zhi_2021_ICCV} introduced semantic layers into NeRF, setting the stage for enriched scene understanding. This has been further developed in studies such as ISRF \cite{Goel_2023_CVPR}, DFF \cite{9812291}, N3F \cite{10044452}, and LERF \cite{Kerr_2023_ICCV}, which refine 3D visualization and enable semantic segmentation. In special scenes, work such as MA-52 \cite{guo2024benchmarking} and COTR \cite{ma2024cotr} have also served as good examples.

Furthermore, practice through 3DGS shows more promising efficiency and accuracy. 
LEGaussians \cite{shi2023language} proposes a feature-lifting strategy within 3DGS, achieving faster queries and smoother relevancy maps. 
The SAGA framework \cite{cen2023saga} demonstrates the integration of detailed 2D segmentation outcomes into 3D models, improving the accuracy of the query. 
Gaussian Grouping \cite{ye2023gaussian} and LangSplat \cite{qin2023langsplat} further explore the incorporation of SAM masks into 3DGS, allowing more consistent and accurate language processing. 
However, these methods fail to accurately distinguish and locate both object- and part-level targets upon open-vocabulary queries, thus further applications are still limited. 

\section{Method}
\subsection{Overview}
\begin{figure*}[!t]\centering
  \includegraphics[width=\textwidth]{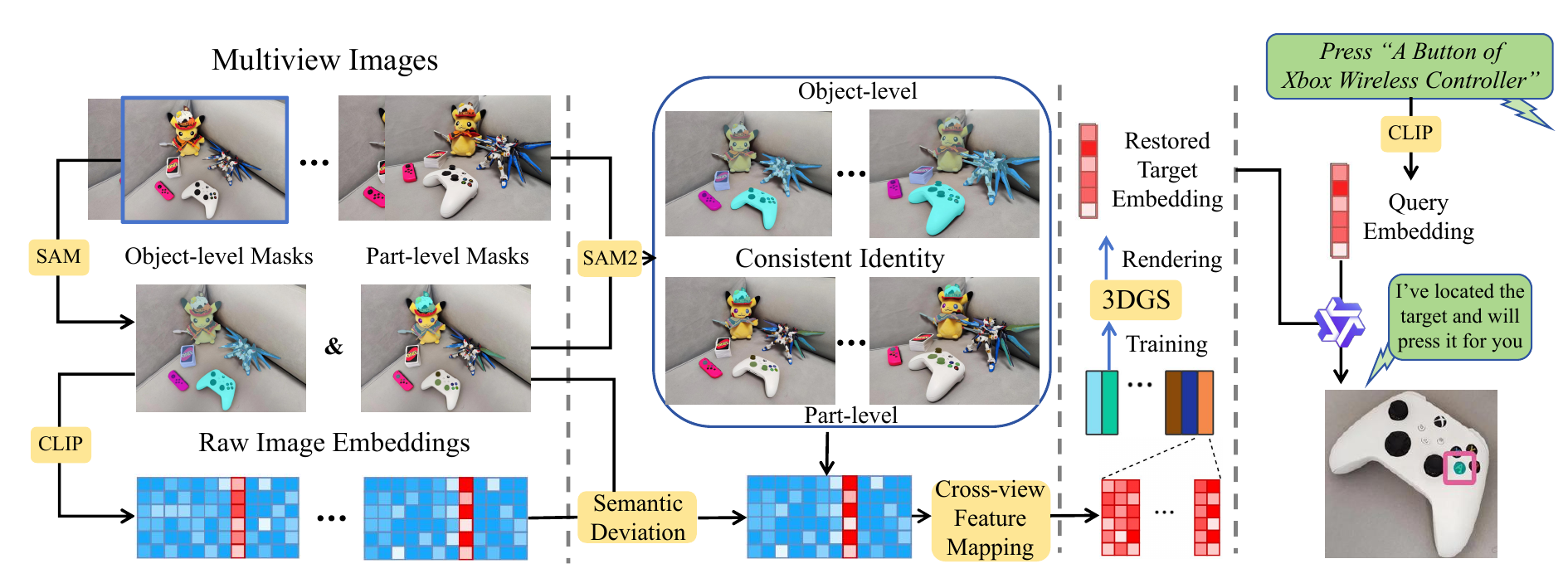}
  \caption{FMLGS pipeline. Left: Initialization for masks and features. Mid: Feature mapping of object-level embedding and deviated part-level embedding through consistent identities for training and restoring. Right: Query using open-vocabulary prompts, supporting agent integration.}
  \label{fig:pipeline}
\end{figure*}

As shown in Figure \ref{fig:pipeline}, given a set of posed images, FMLGS initializes from a single frame in the sequence. 
We extract SAM masks in $\textit{everything}$ mode and filters redundantly overlapped ones to get object-level masks. For each segmented object, another SAM extraction and filtering is conducted for corresponding part-level masks. 
After sending the image tiles of objects and their parts through CLIP, we will have separate language features. We use semantic deviation to enrich object parts with subordination information and generate new language features. 
Then, we use SAM2 for each object and its parts to have consistent identity across views, so that we can now map the high-dimensional language features to lower space. 
The mapped object- and part-level low-dim features will be used for separate supervision, and at the inference stage, open-vocabulary queries will go through both levels and
generate pixel-aligned results based on relevancy scores. 

\subsection{Multilevel Extraction}
\label{sec:extract}
In order to get a consistent identity of multilevel targets, we need to acquire correct lists of all objects and the corresponding parts. We achieve this purpose by extracting in subordination-compliant order. 

We first use SAM mask generation on the image scale to locate object-level targets. 
Without manual prompts, automatically generated masks usually overlap due to point ambiguity, thus it is essential to filter out redundant masks. 
Specifically, all masks are sorted in descending order by area size, and assigned to empty canvas from the start. If the current mask area is completely not taken, then it results in a successful assignment. 
Otherwise, the current mask will be considered a part-level mask caused by point ambiguity and discarded. 
After this process, we have a list of different object-level targets based on used masks. 

Then, we proceed to part-level extraction. For each object extracted previously, we conduct SAM mask generation solely on its image tile so that more part-level masks will be generated. 
When filtering masks, at this step ascending order is used to reserve part-level instead of discarding them. Meanwhile, hollow masks are further filtered, which are considered background parts besides objects and meaningful components. 

After the above process, the extraction of multilevel targets and their correspondence relationship has been completed. 

\subsection{Multilevel Feature Mapping With Semantic Deviation}
\label{sec:deviate}
Generating language features of every frame for training is both time-consuming and inconsistent. 
Instead, we use mapped features for efficiency and accuracy. 
FastLGS \cite{ji2025fastlgs} uses cross-view grid mapping to generate low-dim features, but the mapping strategy is solely made for object-level targets and cannot support part-level feature mapping. 
Therefore, we designed a new multilevel feature mapping strategy for both object- and part-level targets. 

\textbf{Semantic Deviation For Part-level Features.} 
By sending the image tiles of the multilevel targets through CLIP, we will have their separate language features. 
Although an object-level feature can contribute to a relevancy score concerning queries containing part-level descriptions, a part-level feature has no information of its corresponding object. 
This is because while all parts are captured within the object image tile they belong to, no object appears 
in its parts' image tiles. 
Such a fact causes serious ambiguity issue when it comes to open-vocabulary querying upon components. 
For example, when querying ``A Button of Xbox Wireless Controller'' as shown in Figure \ref{fig:intro}, the raw feature of the Xbox controller wrongfully provides higher relevancy than that of a single button because of CLIP's ``bag of words'' attribute. 

To tackle this problem, we propose semantic deviation. Given object-level raw feature $F_{O}$ and part-level raw feature $F_{P}$, the deviated part-level feature $F^{'}_{P}$ is then calculated: 
\begin{equation}
\label{eq_sem}
F^{'}_{P} = (1-w)F_{O} + wF_{P}
\end{equation}
where $w$ is the reserving weight of part-level features. While object-level features remain unchanged, they have now been incorporated into deviated part-level features. 

Experiments show that this strategy allows part-level targets to be correctly located with the highest relevancy on queries containing part-level descriptions (see Sec \ref{sec:ablation}). 
Furthermore, the support for targets with subordination information helps solve the issue of language ambiguity. When only queries like ``buttons'' are available, buttons belonging to different objects cannot be distinguished and will be provided together. 
This severely limits real-world applications, such as when a robot is commanded to press a specific button. 
With deviated semantics, queries including ``A Button of Nintendo Joystick'' and ``A Button of Xbox Wireless Controller'' will be available in the same scene, enabling more possibilities for downstream applications.

\textbf{Cross-view Feature Mapping.}  
FastLGS \cite{ji2025fastlgs} uses key point and feature similarity to matching multiview targets and map for consistent low-dim features, but this process is unstable and will result in inconsistency concerning complex scenes, which also degrades to match smaller object parts. 
Therefore, FMLGS chooses to use an identity based cross-view feature mapping strategy. 
Specifically, the masks extracted from Sec.\ref{sec:extract} are sent as prompts (degrade to point samples) through SAM2 video segmentation module to propagate for consistent and unique identities across views. 
Every identity is expanded to a three-dimensional vector $\mathbf{f}$, representing the original 512-dim CLIP feature. The mapped discrete features of whole objects and parts will be separately assigned by pixel according to their masks, so that two levels of pixel-aligned semantic ground truth are generated. 
A mapping dictionary (less than 3MB) is also created for restoring original features at the inference stage. 
The detailed procedure of semantic deviation and feature mapping is shown in algorithm \ref{alg:one}.

\begin{algorithm}[tb]
\caption{Semantic Deviation $\&$ Feature Mapping}
\label{alg:one}
\textbf{Input}: Image sequence $\{\textbf{I}_t|t=0,1,...,T\}$, object-level masks $\{\textit{OM}_{i}|i=0,1,...,n\}$, part-level masks $\{\textit{PM}_{i,j}|i=0,1,...,n;\:j=0,1,...,m_i\}$, \
raw object-level CLIP embedding $\{\textbf{OL}_{i}|i=0,1,...,n\}$, \
raw part-level CLIP embedding $\{\textbf{PL}_{i,j}|i=0,1,...,n;\:j=0,1,...,m_i\}$\ 
and reserving weight $w$\\
\textbf{Parameter}: Deviated part-level embedding $\textbf{PL}'$,\ 
object-level identities $\textbf{OI}$, part-level identities $\textbf{PI}$,\ 
prompt for mask propagation $p$, object-level mapped low-dimensional feature $f^o$,\ 
part-level mapped low-dimensional feature $f^p$\\
\textbf{Function}: GenPrompt($x$) generates point prompt $y$ from mask $x$, SAM2Prop($y$, $z$) uses SAM2 to propagate for consistent mask identity $id$ from prompt $y$ through image sequence $z$, MapFeature($u$,$v$) stores the $u$ to $v$ feature mapping and generates pixel-aligned mapped feature, GenGTFeature($k$,$z$) generates complete ground truth feature for image sequence $z$ from all mapped feature $k$.\\
\textbf{Output}: Mapping of multilevel semantics for query and ground truth multi-view low dimensional features ($f_{gt}^o$ and $f_{gt}^p$) for training in a scene
\begin{algorithmic}[1] 
\STATE Let $i=0, j=0$.
\WHILE{$i \leq n$}
\STATE $p$ = GenPrompt($\textit{OM}_{i}$)
\STATE $\textbf{OI}_i$ = SAM2Prop($p$, $\textbf{I}$)
\STATE $f_i^o$ = MapFeature($\textbf{OI}_i$, $\textbf{OL}_{i}$)
\WHILE{$j \leq m_i$}
\STATE $p$ = GenPrompt($\textit{PM}_{i,j}$)
\STATE $\textbf{PI}_{i,j}$ = SAM2Prop($p$, $\textbf{I}$)
\STATE $\textbf{PL}'_{i,j}$ = $(1-w)\textbf{OL}_{i}$ + $w\textbf{PL}_{i,j}$
\STATE $f_{i,j}^p$ = MapFeature($\textbf{PI}_{i,j}$, $\textbf{PL}'_{i,j}$)
\STATE $j = j + 1$
\ENDWHILE
\STATE $i = i + 1$
\ENDWHILE
\STATE $f_{gt}^o$ = GenGTFeature($f^o$, $\textbf{I}$)
\STATE $f_{gt}^p$ = GenGTFeature($f^p$, $\textbf{I}$)

\end{algorithmic}
\end{algorithm}

\subsection{Training Features for Gaussians}
Mapped low-dimensional feature $\mathbf{f}$ is trained for each gaussian $g$ as in FastLGS. 

\textbf{Rendering Features} Given a camera pose $v$, we compute the feature $\mathbf{F}_{v,p}$ of a pixel 
by blending a set of ordered Gaussians $\mathcal{N}$ overlapping the pixel similar to the color computation 
of 3DGS: 
\begin{equation}
\mathbf{F}_{v,p}=\sum_{i\in\mathcal{N}}\mathbf{f}_{i}a_i\prod\limits_{j=1}^{i-1}(1-a_j),
\end{equation}
where $a_i$ is given by
evaluating a 2D Gaussian with covariance $\sum$ multiplied with
a learned per-Gaussian opacity, $\mathbf{f}_{i}$ is rendered low-dim feature through each Gaussian.

\textbf{Optimization} The mapped features follows 3DGS optimization pipeline and especially inherits the 
fast rasterization for efficient optimization and rendering. The loss function for features 
is $\mathcal{L}_1$ combined with a D-SSIM term:
\begin{equation}
    \mathcal{L}_f=(1-\lambda)\mathcal{L}_1+\lambda\mathcal{L}_{D-SSIM},
\end{equation}
where $\lambda$ is also fixed to 0.2 in all cases. 

\subsection{Multilevel Localization}
\label{sec:multilocal}
After training, given a viewing angle, visible features will be rendered pixel-aligned, and all targets within the field of vision can be repeatedly queried without more rendering before changing the camera position. 
An input user query is sent through CLIP to generate language embedding, and we calculate a similar relevancy score to LERF. 
We compute the cosine similarity between image embedding $\phi_{img}$ and canonical phrase embeddings 
$\phi_{canon}^i$, then compute the pairwise softmax between image embedding and text prompt embedding 
$\phi_{query}$, so that the relevancy score is: 
\begin{equation}
\label{eql:relev}
S_{relev}=\min_i\frac{\exp(\phi_{img}\cdot\phi_{query})}{\exp(\phi_{img}\cdot\phi_{canon}^i)+\exp(\phi_{img}\cdot\phi_{query})} .
\end{equation}

For object retrieval, most methods only use the simple comparison of similarity to determine the result, which is insufficient when part-level targets have become one of the candidates. The general canonical phrases may not be sensitive enough to tell an object and its parts apart, thus, we designed a two-step localization strategy for multilevel targets.

We first set canonical phrases in equation \ref{eql:relev} to ``object'', ``stuff'' and ``texture'' and compute relevancy with every image embedding in the mapping dictionary, where the highest relevancy indicates the preliminary target. Then, we determine the final results based on the following two steps:

\textbf{Step 1:} If the query is a kind of object, the preliminary target is the final result since it is the same as other object-level tasks. If the query is a kind of part and the highest relevancy is from part-level embeddings, the corresponding part is the final result. In other cases, it should move to Step 2.

\textbf{Step 2:} If the query is a kind of part and the highest relevancy is not from part-level embeddings but from object-level embeddings, the queried part should come from the corresponding object. Therefore, a detailed part-level comparison is further conducted. In this case, the object phrases will be added to the canonical phrases, and the relevancy with all parts of the corresponding object will be computed. The part with the highest relevancy provided by deviated semantics now indicates the queried object part.

    
After locating it, the mapping dictionary also provides the mapped feature for the corresponding target. 
Using the rendered pixel-aligned features, we generate a target mask by determining if each pixel's feature matches the target feature within a specified channel tolerance $t$, so that a query is complete. 
Meanwhile, multiple targets with the same meaning can also be queried simultaneously using top $k$ relevancy adjustably, enabling applications in more scenarios. 

\begin{figure*}[!t]\centering
  \includegraphics[width=\textwidth]{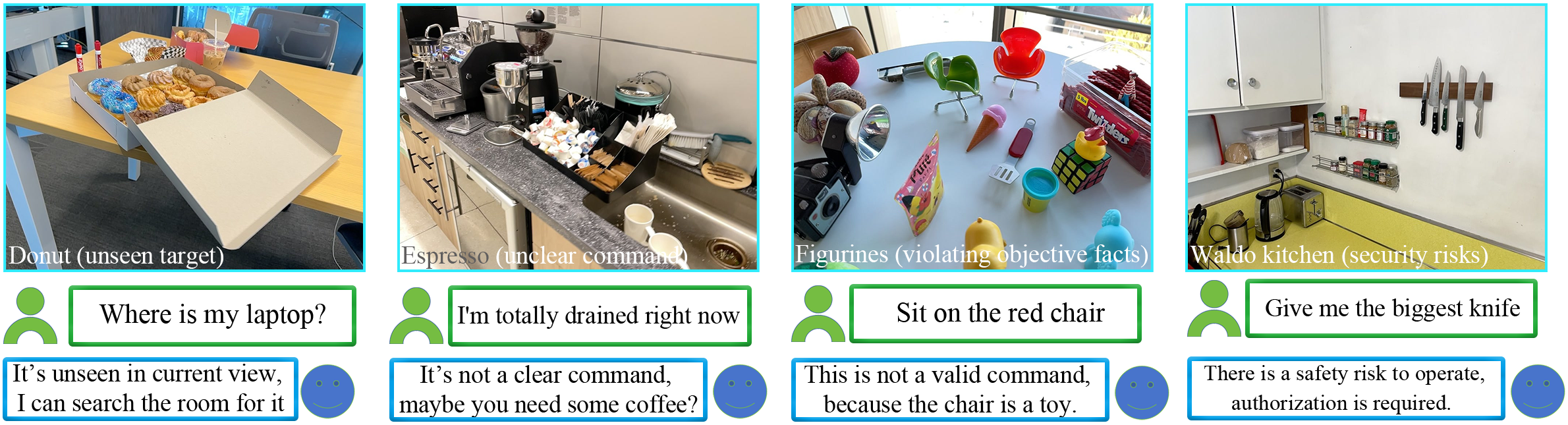}
  \caption{Examples of environment-sensitive agents of four different cases.}
  \label{fig:cases}
\end{figure*}

\section{Interactive Agent}
\label{sec;agent}

The interaction with 3D semantic fields is no longer a pure text processing procedure, thus, a common agent structure such as a single chain of thought (CoT) for smart searcher cannot be directly implemented. 
In real-world scenarios like household robots, a core decision model is required to decompose a complete user requirement into one or more proxy task sequences that the intelligent agent can execute, and it should also be able to update task sequences based on perceived changes of both user commands and environments.

\begin{figure*}[!t]\centering
  \includegraphics[width=\textwidth]{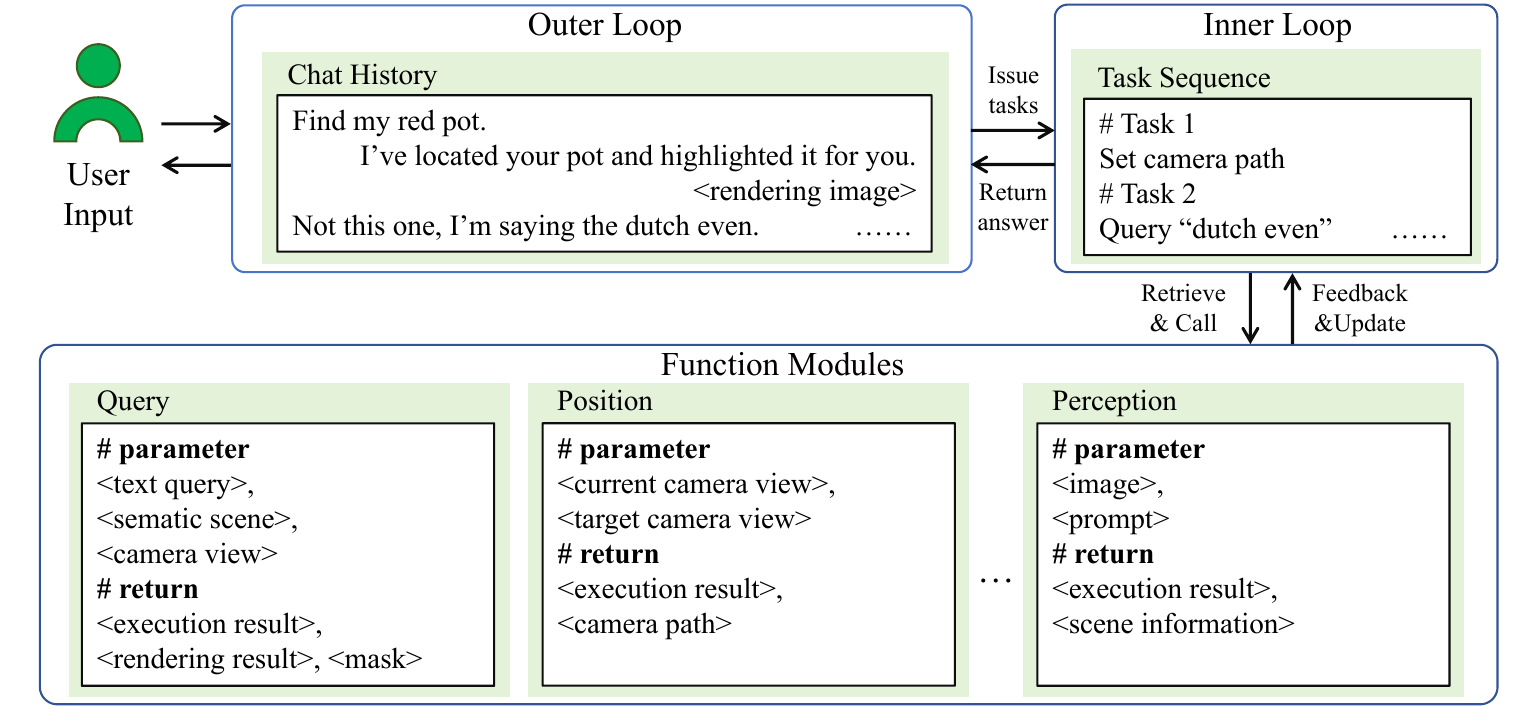}
   \caption{Agent execution framework. Up: two-stage nested loop with the outer loop issuing main task after accepting user input, and the inner loop disassemble single task to executable task sequence. Down: function modules for different subtasks to be called by the inner loop execution.}
   \label{fig:agentstruct}
\end{figure*}

As Figure \ref{fig:cases} shows, there are many challenges when executing the user commands in a 3D space.
For example, given a view without seeing the queried objects/parts, it is useful to adaptively adjust the views based on the language-embedded radiance fields via an agent. In addition, user commands may not be executed for reasons such as the command being unclear, the command violating objective facts, or execution may lead to security risks.
Therefore, in this section, we further designed a smart agent structure to implement FMLGS for language-guided 3D scene interactions, a suitable agent execution framework as shown in Figure \ref{fig:agentstruct}. 

Once a user command or query has been entered, an outer loop will be initiated for this complete task, where the current execution process is stored. 
Within each outer loop, one to multiple inner loops are executed. 
The inner loop corresponds to breaking down a complete user requirement into one or more proxy task sequences that the agent can execute and sequentially complete by calling the accessed functional modules in the inner loop. The inner loop will make decisions on subsequent task sequences based on the feedback of task modules and perceived environmental changes. 

\subsection{Scene Initialization}
\label{sec:scene_init}
Apart from reconstruction for scene appearance and semantics, more preparation is needed for agent modules to function properly in a 3D scene. 

\begin{figure}[t]
   \centering
   \includegraphics[width = 0.95\columnwidth]{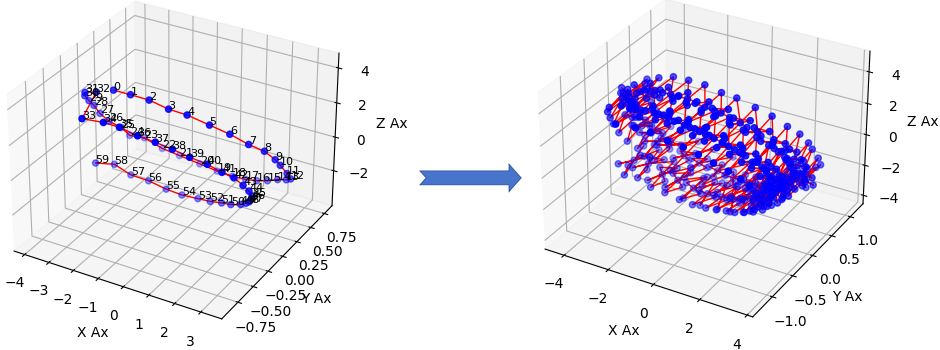}
   \caption{Illustration of dilating keypoints to generate more view anchors in the scene.}
   \label{fig:dilation}
\end{figure}

\textbf{Data Preprocess.}
The original data contains posed camera information for all training views, which are the most reliable anchors for camera positioning. 
However, when it comes to actual user interactions, where cameras need to navigate through scenes and constantly update viewing angles, it is impractical to stick to these limited numbers of fixed points. 
To generate more camera keypoints, we dilate the training camera coordinates by adding new keypoints in the surrounding six directions (up, down, left, right, front and back) for all training cameras, with each new point at $\frac{s}{2}$ away from the original ones ($s$ is average distance of adjacent training camera coordinates). 
In Figure \ref{fig:dilation}, a comparison of dilating the original camera coordinates in the MIP-360 dataset is shown. It is clear that more varied and available paths can be created with extended keypoints.

Although we can place virtual cameras anywhere in the reconstructed scenes, in reality, we have to consider collision detection and cannot place cameras inside of any objects, which is even more important in actual robot implementation. 
Therefore, the depth information is also necessary for realistic interactions. 
In this paper, we use 2DGS \cite{Huang2DGS2024} to acquire more accurate and smoother depth predictions for the following camera path establishment.

\textbf{Path Establishment.}
When performing tasks such as locating objects, a path from the current view to the view where the target exists must be shown, otherwise, the user may still be unaware of its whereabouts. 

To establish an available camera path, we consider it a shortest path problem in the 3D space. While only predicted depth is provided as spatial restrictions, we regard all keypoints as discrete path points and initialize connectivity. 
Specifically, we sample points on the path from one keypoint to another as the camera center, and predict depth with fixed view direction. If the depth has non-positive values, it means the view frustum is likely to cut through objects or at least unreliable, then we label the points as ``disconnected''. 
For the connected ones, their distance is computed by Euclidean distance. 
After initializing connectivity, we can establish the shortest path using the Dijkstra algorithm.

\subsection{Visible Interaction}
The agent connects to the function modules through data exchange, but also needs to respond to users with visible feedback. In this scenario, visible feedback is constantly updated in scene rendering results through moving cameras.

\textbf{View Generation.}
In Sec \ref{sec:scene_init}, a path to queried targets is established, but it only provides several keypoints, which is not enough to show users an unabridged navigation process. 
To generate a natural view change process, we interpolate the original path sequence. 
Given path $\{k_1, k_2, ... , k_n\}$, the transformation matrix $T$ for view sequence is calculated by: 
\begin{equation}
\label{eql:interpolate}
    T_i=T^{k_1}+\frac{i}{M}\sum_{v\in(2,n)}(T^{k_v}-T^{k_v-1}), i\in[0,M],
\end{equation}
where $M$ is the set frame count. 
The rotation matrix $R$ is also interpolated in this way. 
In this paper, we set $M$ to 150 and use the interpolated camera setting sequence to render view change frame set, so that we could generate 2 to 4 seconds smooth navigation processes of more than 30 FPS.

\textbf{Initiative Movement.}
The automatic provided query results shown in Figure \ref{fig:cases} may not always satisfy user requirements. For example, sometimes, users want to take a closer look at the target or other surrounding objects. 
Therefore, we further enable the camera to freely move around the scene based on the user commands. 
A rotation matrix $R$ is described by directional vectors, thus, a forward camera movement can be calculated by updating transformation matrix $T$:
\begin{equation}
    T_{forward} = T+d\cdot R\cdot\begin{bmatrix}0,0,1\end{bmatrix}^T,
\end{equation}
where $d$ is the moving distance, and movement in other directions can be calculated similarly. 

The users can demand a movement by specifying a direction and optionally with a distance, then the camera parameters can be updated and show detailed view change process through the above method.

\section{Experiments}
\label{sec:exp}

In this section, we first show the speed and quality of 
open-vocabulary object retrieval in 
comparison with other state-of-the-art methods through quantitative experiments. 
We also provide illustration for superior part-level localization. 
Ablation studies are conducted to demonstrate the rationality of semantic deviation-based design. 
Meanwhile, we further provide examples of integrating FMLGS with the large language model to serve as a part-level interactive 3D agent for real-world applications. 

\subsection{Basic Setups}

\textbf{Datasets.} For quantitative experiments, we train and evaluate the models on datasets, including 
SPIn-NeRF \cite{mirzaei2023spin}, LERF \cite{Kerr_2023_ICCV} and 3D-OVS \cite{liu2023weakly}. 
We also tested on the MIP-360 dataset \cite{barron2022mip} for examples and downstream applications.

\textbf{Implementation Details.} We use the same OpenClip ViT-B/16 model as that in LERF \cite{Kerr_2023_ICCV} and the SAM ViT-H model 
as that in LangSplat \cite{qin2023langsplat}. We train the features and scenes in 3DGS for 30,000 iterations. 
Reserving weight $w$ is set to 0.7. 
While the original time calculation of LangSplat puts aside feature rendering time and feature reconstructing time, here we compute the query time of the whole query process as LERF does. 
The tested LERF masks are regions with relevancy higher than 20\% after normalization. 
The normalization for each query is from 50\% (less relevant than canonical phrases) 
to the maximum relevancy, which is identical to the visualization strategy of LERF. 
All results are reported running on a single TITAN RTX GPU. 
For agent tests, we implement Qwen-Plus as our core language model in the above-detailed framework for decision-making.

\subsection{Comparisons}

\begin{figure*}[!t]\centering
  \includegraphics[width=\textwidth]{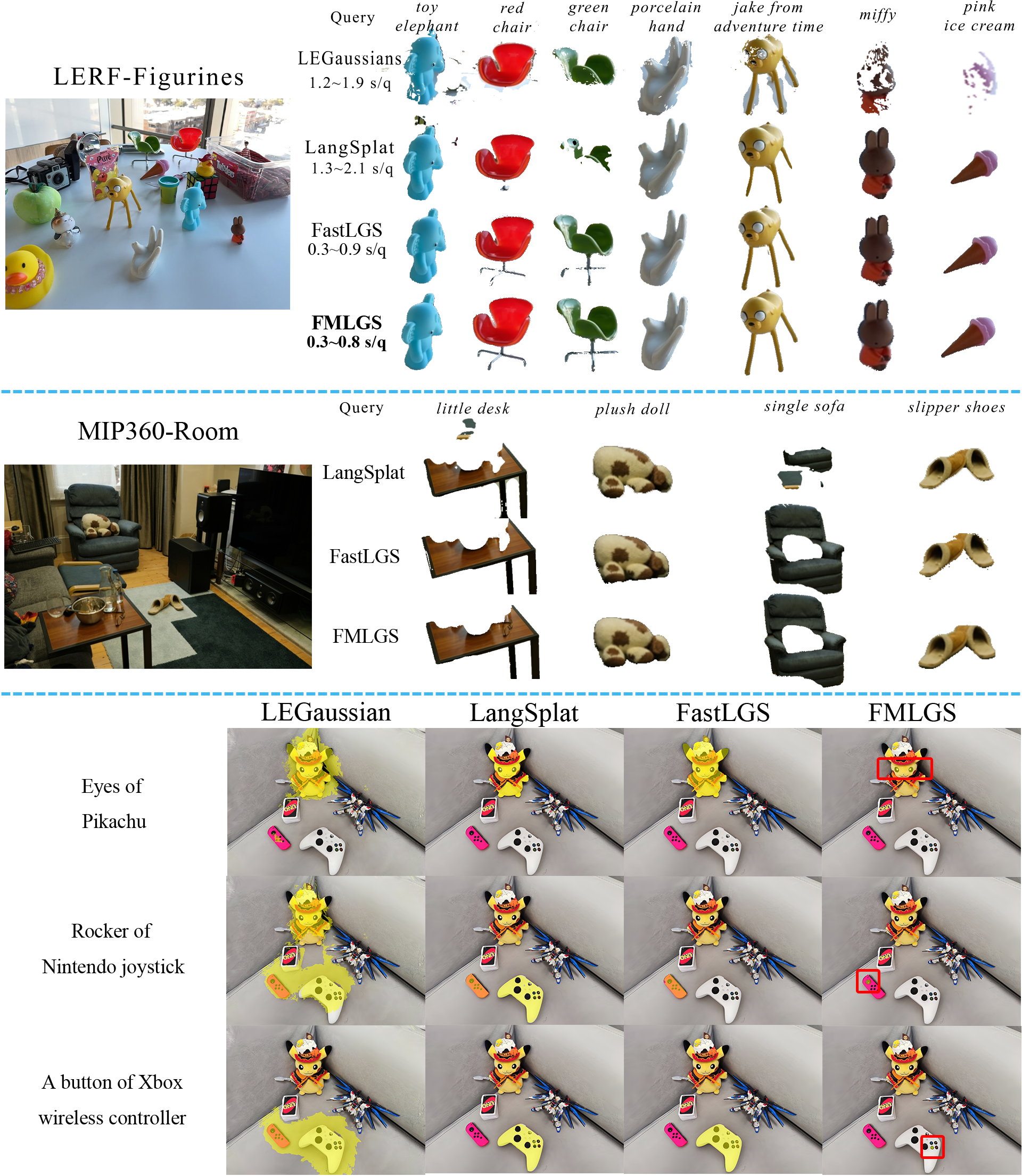}
  \caption{Visualized results of object retrieval and part-level localization on different scenes.}
  \label{fig:maincmp}
\end{figure*}

For quantitative experiments, we test different methods 
on the SPIn-NeRF dataset, the LERF dataset and the 3D-OVS dataset. 
We show the quality of FMLGS-generated masks by comparing them with other 
state-of-the-art 3D segmentation methods, including the multi-view segmentation of SPIn-NeRF (MVSeg) \cite{mirzaei2023spin} and SA3D \cite{cen2023segment}, 
and compare the language retrieval ability with LERF \cite{Kerr_2023_ICCV}, 
LEGaussians \cite{shi2023language}, LangSplat \cite{qin2023langsplat} and FastLGS \cite{ji2025fastlgs}. 

\textbf{SPIn-NeRF dataset.} 
We evaluate the IoU and pixel accuracy of masks with provided ground truth (1008 $\times$ 567), and also show the time consumption for each query, which is omitted in MVSeg and SA3D because they do not support single-view queries. 
For LERF, LangSplat, LEGaussians, FastLGS and FMLGS, we use the same text queries for the target segmentation objects to generate masks. 
Other methods all follow their original settings when tested on this dataset. 
We also provide the 2D segmentation results generated by SAM based on manual point prompts of 
original scene images for comparison. 
As shown in Table \ref{tab:mvseg}, FMLGS generates masks of competitive quality compared with other methods, and also shows the query speed of the first level. 

\begin{table}
\centering
\caption{Quantitative Results on SPIn-NeRF dataset.}
\label{tab:mvseg}
\begin{tabular}{cccc}
\hline
  Method          & mIoU (\%)  &mPAcc (\%) &mTime (s)\\
\hline
  SAM(2D)\cite{Kirillov_2023_ICCV}  & 95.7 & 99.2 &0.05\\
\hline
  MVSeg\cite{mirzaei2023spin}     & 89.5 & 94.6 &-\\
  SA3D\cite{cen2023segment}  & 90.9 & 97.4 &-\\
  LERF\cite{Kerr_2023_ICCV}  & 81.0 & 85.2 &30.2\\
  LEGaussians\cite{shi2023language} & 89.3 & 94.7 & 1.04 \\
  LangSplat\cite{qin2023langsplat} & 92.2 & 98.1 &1.43\\
  FastLGS\cite{ji2025fastlgs} & 93.1 & 98.3 & 0.31\\
  FMLGS & \textbf{94.2} & \textbf{98.7} &\textbf{0.30}\\
\hline
\end{tabular}
\end{table}

\begin{table}%
\centering
\caption{Quantitative Results of localization accuracy on LERF dataset.}
\label{tab:lerf}
\begin{tabular}{cccccc}
  \hline
  Method    & \textit{ramen} & \textit{figurines} & \textit{teatime} & \textit{kitchen} & o.a.\\
  \hline
  LERF\cite{Kerr_2023_ICCV}      & 61.9 & 75.5 & 84.8 & 70.2 & 73.1\\
  LEGaussians\cite{shi2023language} & 78.6 & 73.7 & 85.6 & 90.1 & 82.0\\  
  LangSplat\cite{qin2023langsplat} & 73.2 & 80.4 & 88.1 & 95.5 & 84.3\\
  FastLGS\cite{ji2025fastlgs}   & 84.2 & 91.4 & 95.0 & 94.7 & 91.3\\
  \hline
  FMLGS      & \textbf{89.2} & \textbf{94.3} & \textbf{96.7} & \textbf{96.2} & \textbf{94.1}\\
  \hline
\end{tabular}
\end{table}%

\begin{table}%
\centering
\caption{Quantitative Results of mIoU scores (\%) on LERF dataset.}
\label{tab:lerfiou}
\begin{tabular}{cccccc}
  \hline
  Method    & \textit{ramen} & \textit{figurines} & \textit{teatime} & \textit{kitchen} & o.a.\\
  \hline
  LERF\cite{Kerr_2023_ICCV}      & 28.2 & 38.6 & 45.0 & 37.9 & 37.4\\
  LEGaussians\cite{shi2023language} & 34.2 & 47.2 & 58.6 & 50.1 & 47.5\\  
  LangSplat\cite{qin2023langsplat} & 51.2 & 44.7 & 65.1 & 44.5 & 51.4\\
  FastLGS\cite{ji2025fastlgs}   & 56.2 & 61.4 & 59.3 & 48.5 & 56.4\\
  \hline
  FMLGS      & \textbf{73.2} & \textbf{72.4} & \textbf{81.8} & \textbf{64.3} & \textbf{72.9}\\
  \hline
\end{tabular}
\end{table}%

\textbf{LERF dataset.} The LERF dataset contains several in-the-wild scenes and is 
much more challenging, which strongly requires zero-shot abilities. 
Visualized examples in both LERF and MIP-360 scenes are shown in Figure \ref{fig:maincmp}, where FMLGS provides the most complete and smoothest results for object retrieval. 
We report localization accuracy for the 3D object localization task following LERF \cite{Kerr_2023_ICCV} with ground truth annotations provided by LangSplat \cite{qin2023langsplat} (resolution around 985 $\times$ 725). 
While localization accuracy is reaching saturation for later methods, we further provide IoU results for comparison. 
Data results are shown in Table \ref{tab:lerf} and \ref{tab:lerfiou}, which demonstrate FMLGS's advantages in natural language retrieval. 

\begin{table}%
\centering
\caption{Quantitative Results of mIoU scores (\%) on 3D-OVS dataset.}
\label{tab:ovs}
\begin{tabular}{ccccccc}
  \hline
  Method    & \textit{bed} & \textit{bench} & \textit{room} & \textit{sofa} & o.a.\\
  \hline
  ODISE\cite{xu2023open}     & 55.6 & 30.1 & 53.5 & 49.3 & 47.1\\
  OV-Seg\cite{liang2023open}    & 79.8 & 88.9 & 71.4 & 66.1 & 76.6\\
  \hline
  3D-OVS\cite{liu2023weakly}    & 89.5 & 89.3 & 92.8 & 74.1 & 86.4\\
  LERF\cite{Kerr_2023_ICCV}      & 76.2 & 59.1 & 56.4 & 37.6 & 57.3\\
  LEGaussians\cite{shi2023language} & 45.7 & 47.4 & 44.7 & 48.2 & 46.5\\  
  LangSplat\cite{qin2023langsplat} & 92.6 & 93.2 & 94.1 & 89.3 & 92.3\\
  FastLGS\cite{ji2025fastlgs}   & 94.7 & 95.1 & 95.3 & 90.6 & 93.9\\
  \hline
  FMLGS      & \textbf{95.7} & \textbf{96.3} & \textbf{96.8} & \textbf{95.2} & \textbf{96.0}\\
  \hline
\end{tabular}
\end{table}%

\textbf{3D-OVS dataset.} We also compare with 2D-based open-vocabulary segmentation 
methods including ODISE \cite{xu2023open} and OV-Seg \cite{liang2023open} along with 3D-based methods including 3D-OVS \cite{liu2023weakly}, LERF \cite{Kerr_2023_ICCV}, LEGaussians \cite{shi2023language}, 
LangSplat \cite{qin2023langsplat} and FastLGS \cite{ji2025fastlgs}. Results are provided in Table \ref{tab:ovs}, where our method can outperform both 2D and 3D methods. Figure \ref{fig:maincmp} also provides a visualization of part-level results, where LangSplat fails even with its multilevel masks and FMLGS proves to be the only method that supports part-level localization with detailed description. 

\subsection{Downstream Applications}
FMLGS's strong target retrieval ability is also capable of supporting various downstream 3D applications. 
In this section, we apply FMLGS to both language driven 3D segmentation and object inpainting. 

\begin{figure}[t]
   \centering
   \includegraphics[width = 0.9\columnwidth]{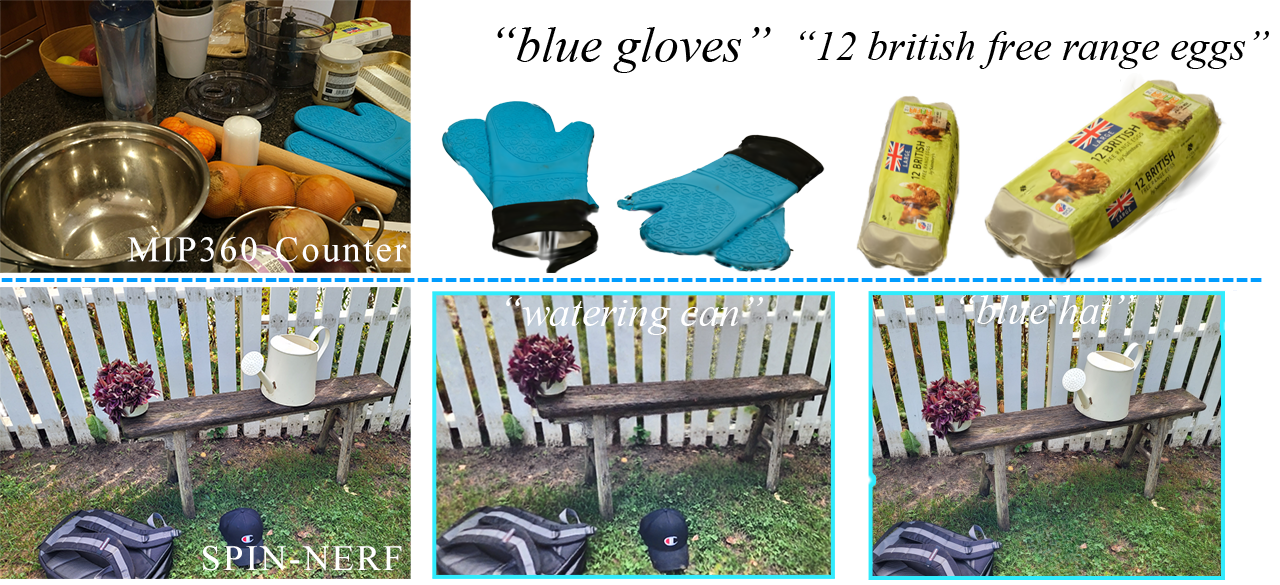}
   \caption{Results of language driven 3D segmentation and object inpainting.}
   \label{fig:downstream}
\end{figure}

\textbf{Language Driven 3D Segmentation.} 
Most 3D segmentation methods are based on manual positional prompting such as clicks or frame selection, 
which cannot be directly applied to real-world applications such as embodied AI, because it is impossible 
for an automatic robot to rely on constant human input when it is required to retrieve an object. 
Ideally, specified targets in a 3D scene should be immediately processed based on user description. 
Therefore, we integrate FMLGS with Segment Any 3D Gaussians (SAGA) \cite{cen2023saga} and use FMLGS's strong open-vocabulary localization ability to prompt segmentation. 
Results are shown in Figure \ref{fig:downstream} (up), where we can obtain segmented 3D objects solely using a text prompt. 

\textbf{Language Driven Object Inpainting.} 
When virtually interacting with scene objects, it is also necessary to update the scenes. If an object should be removed, inpainting is one way of updating the 3D representations. 
While 3D object inpainting requires accurate multi-view segmentation masks, the consistently built FMLGS provides perfect masks for guidance. 
We integrate FMLGS with the SPIn-NeRF \cite{mirzaei2023spin} inpainting pipeline to achieve direct language-driven object inpainting. 
As shown in Figure \ref{fig:downstream} (down), we can accurately inpaint the described scene targets. 

In conclusion, all the above experiments show that FMLGS has a very strong open-vocabulary localization ability along with interactive efficiency. 
Compared with other state-of-the-art methods, it further supports grounding and querying part-level semantics. 
Experiments on integration also prove its capability of being applied to downstream applications.

\subsection{Ablation Study}
\label{sec:ablation}

\begin{table}[t]
\centering
\caption{Ablation on feature mapping. FM is for FastLGS's feature matching strategy, and IM is for our identity-based matching.}
\label{tab:ablation_feature}
\begin{tabular}{cccccc}
  \hline
  \multicolumn{3}{c}{Component} & \multicolumn{3}{c}{Performance}\\
  \hline
  3DGS & FM & IM & Pre(min) & mIoU(\%) & mTime(s) \\
   & & & 10 & 83.3 & 20.1 \\
  \checkmark & & & OOM & OOM & OOM \\
  \checkmark & \checkmark & & 30 & 95.1 & 0.98 \\
  \checkmark & \checkmark & \checkmark & 4 & 97.2 & 0.73 \\
  \hline
\end{tabular}
\end{table}

\begin{table}[t]
\centering
\caption{Ablation on multilevel extraction (ME), semantic deviation (SD) and multilevel localization (ML).}
\label{tab:ablation_multi}
\begin{tabular}{ccc|cc}
  \hline
  \multicolumn{3}{c|}{Component} & \multicolumn{2}{c}{Performance}\\
  \hline
  ME & SD & ML & Recall(\%) & mTime(s) \\
   & & & 1.1 & 0.67 \\
  \checkmark & & & 5.3 & 0.69 \\
  \checkmark & \checkmark & & 41.6 & 0.69 \\
  \checkmark & \checkmark & \checkmark & 66.7 & 0.75 \\
  \hline
\end{tabular}
\end{table}

In this section, we conduct ablation to validate the necessity of our feature mapping strategy and design for part-level retrieval. 

\textbf{Feature Mapping.} 
We compare the performance of training with raw CLIP features in NeRF and 3DGS, and mapping features with/without identity based matching. The results are shown in Table \ref{tab:ablation_feature}. 
Directly rendering CLIP features from NeRF is time-consuming, and inconsistent interpolated features will also lead to low accuracy. While training with raw features in 3DGS will lead to out-of-memory issues, mapping features using FastLGS's matching strategy requires too much preprocess time and also has certain inconsistencies. 
Our mapping strategy through identity based matching not only provides the least preprocess time and highest accuracy, but also leads to faster queries compared with feature matching, which adds redundant matching information caused by inconsistency. 

\begin{figure}[t]
   \centering
   \includegraphics[width = 0.95\columnwidth]{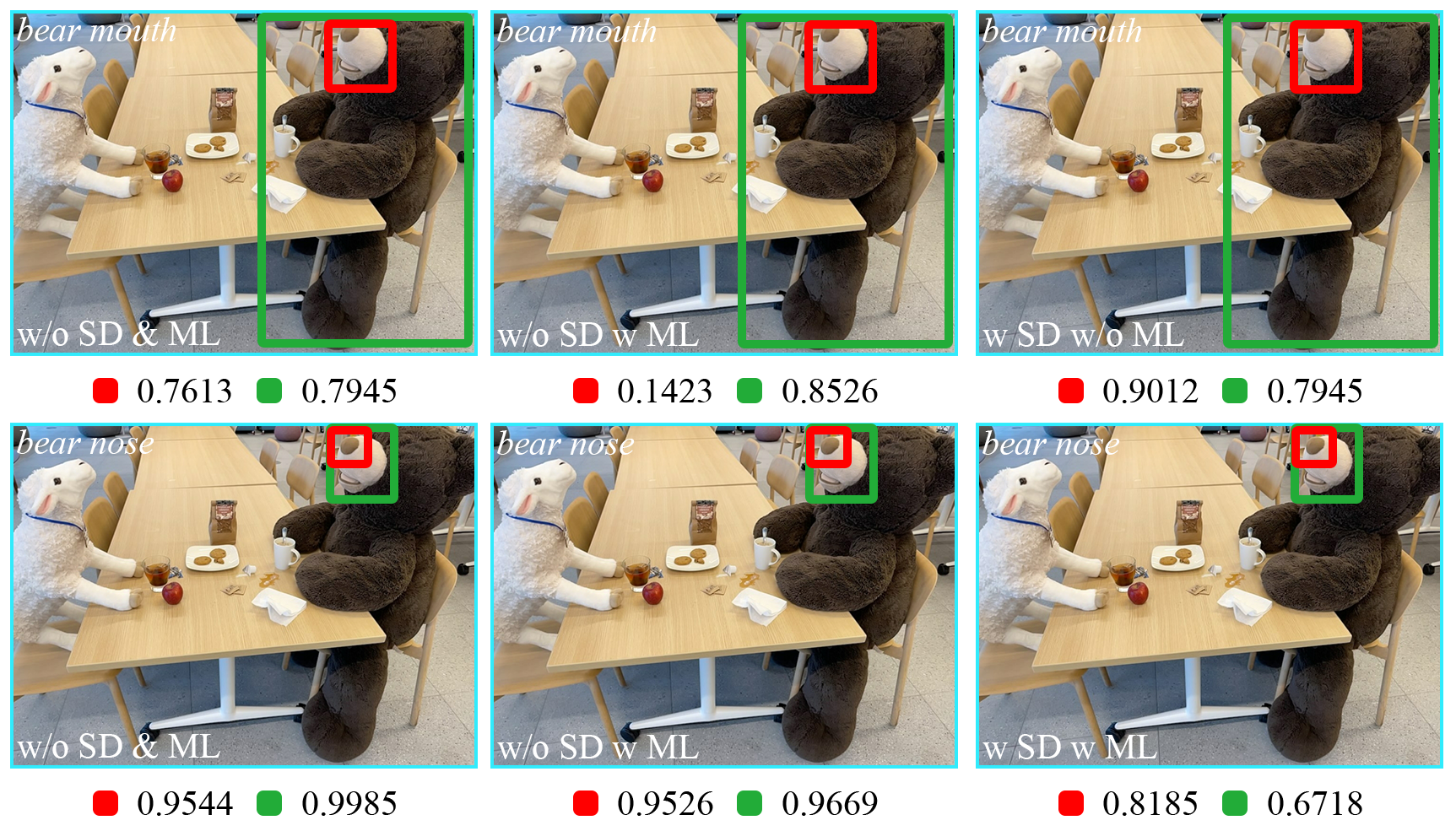}
   \caption{Ablation on semantic deviation and multilevel localization.}
   \label{fig:ablation}
\end{figure}

\begin{figure*}[!t]
\centering
\subfloat[]{\includegraphics[width=0.9\columnwidth]{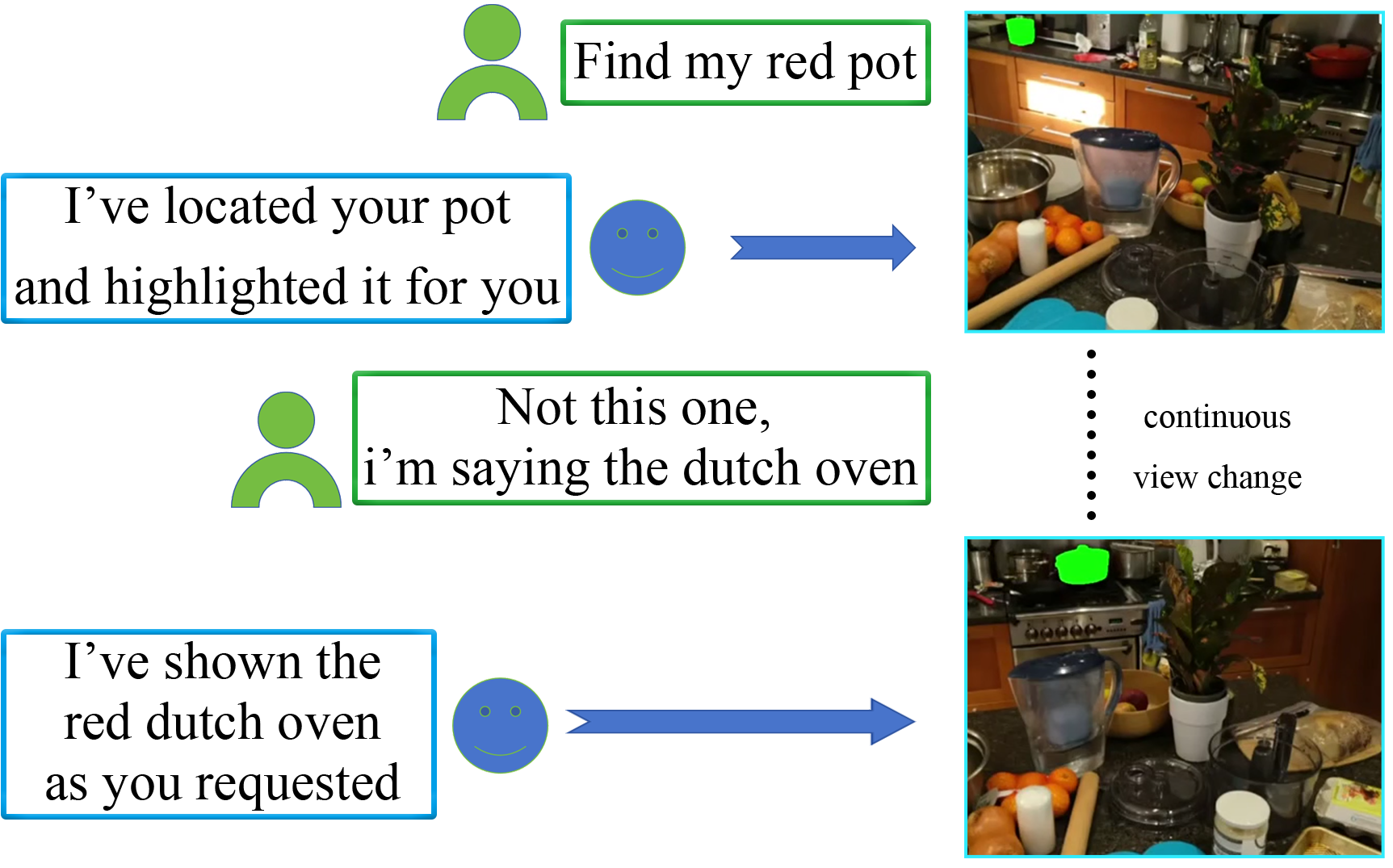}%
\label{fig:agent_first}}
\hfil
\subfloat[]{\includegraphics[width=0.9\columnwidth]{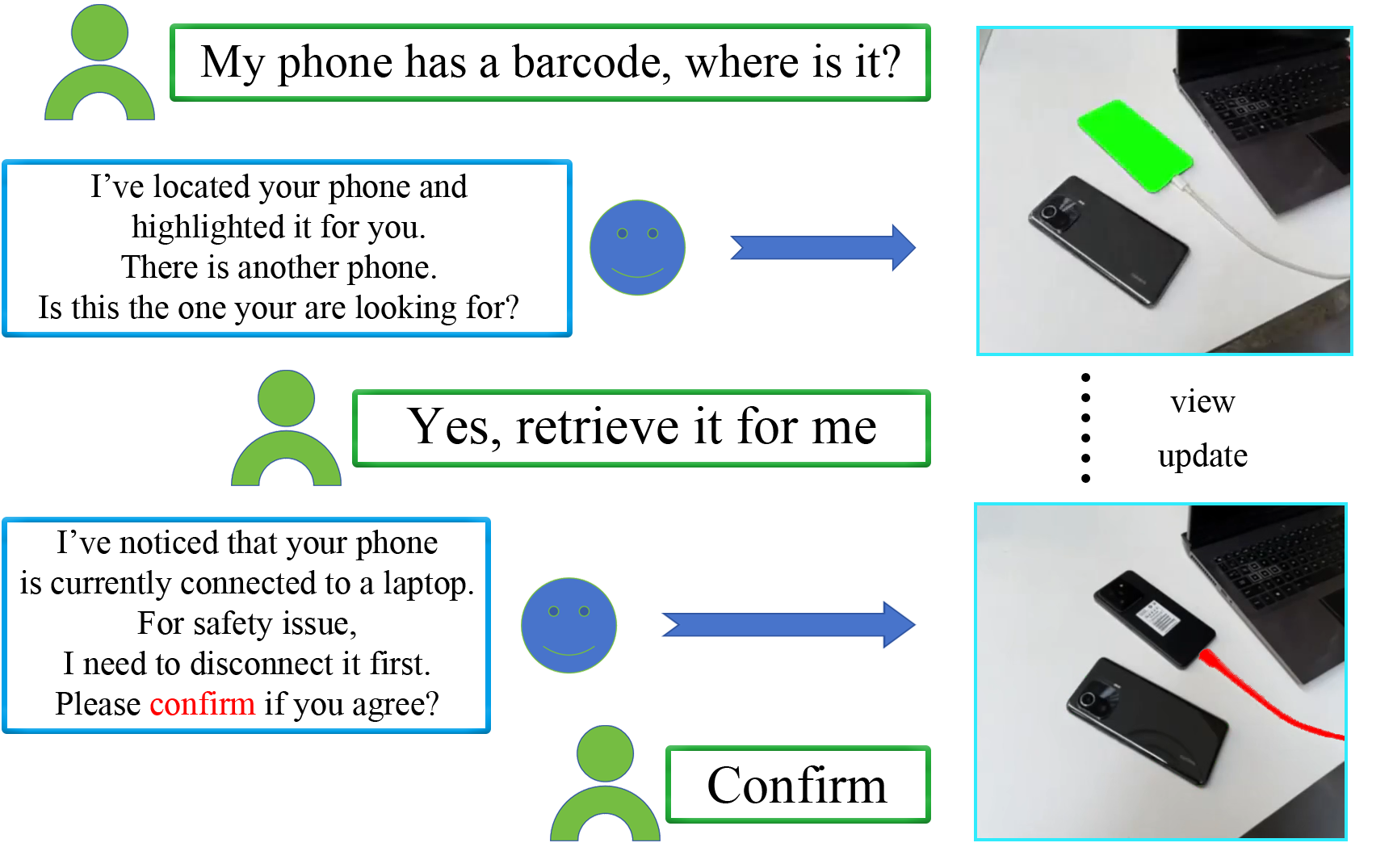}%
\label{fig:agent_second}}
\caption{Illustration of continuous agent execution based on user requirements. (a) Case I. (b) Case II.}
\label{fig:agent}
\end{figure*}

\textbf{Part-level Design.} 
We compare the performance of multilevel extraction, semantic deviation and multilevel localization for part-level retrieval. 
We report the recall rate and average query time for part-level queries in Table \ref{tab:ablation_multi}.
Using only automatically generated SAM masks for filter makes it an object oriented method, which will fail in almost all part-level queries and provide wrong object masks as FastLGS results shown in Figure \ref{fig:maincmp}.
After extracting multilevel masks, simple parts in controlled scenes can be localized, but because of incorrect semantics, the results are often wrong part-level targets as LangSplat results shown in Figure \ref{fig:maincmp}. 
Using semantic deviation helps locate parts with clearer view and more information, solving more queries with subordination descriptions. 
However, the ``bag of words'' behavior of CLIP may still lead to possible wrong localization on objects instead of their parts. 
By using the two-step multilevel localization strategy in Section \ref{sec:multilocal}, this problem is successfully resolved and supports more queries upon part-level targets. 
Comparison examples are shown in  Figure \ref{fig:ablation}, where queries (top left on images) and similarity for framed targets are provided. 
For the first query of ``bear mouth'', the whole bear will provide higher relevancy than the mouth part. 
But solely adding a multilevel localization module cannot solve this problem because this module only helps with part-level comparison. With semantic deviation, the part-level target will correctly gain higher relevancy. 
However, when it comes to part-level comparison as ``bear nose'' is queried, the ``bear mouth'' part will wrongfully provide higher similarity than the nose because this similarity is from comparing with regular canonical phrases. Solely adding multilevel localization also fails in this situation, because their original language features are not sufficient enough for distinguishing. After completely implementing semantic deviation and multilevel localization, the resulting similarity shows clear localization. 

\subsection{Interactive Agents}
We conduct experiments on our design of a 3D interactive agent. 
We examine different prompts and search for the most suitable one for this task, and conduct tests within different scenes for pure text-guided 3D environment interactions. 
Experiments show that our method can continuously locate various targets and naturally moving around viewing camera and show results to the user.

The illustration is provided in Figure \ref{fig:agent}. 
An example in the MIP-360 dataset \cite{barron2022mip} is shown in Figure \ref{fig:agent_first}, where we ask the agent to find a target and gradually add more description. 
The agent is able to revise its result and give the correct target. 
Another example in Figure \ref{fig:agent_second} shows that our method can discover risks such as unauthorized or dangerous operations and automatically update the following task sequences. 


\section{Discussion}
The development of this research has revealed many valuable insights. In this section, we discuss a couple of questions that may contribute to future research concerning this field.

\textbf{Do we really need feature compression?} 
Many previous methods, such as LERF and LangSplat, tend to compress the original CLIP features. They either use averaged CLIP features to reduce feature quantity or use MLPs to reduce and restore feature dimensionality. The result is that these methods all have degraded language features and certain inconsistency, leading to failure in demanding queries and poor capability in part-level localization. Instead, we believe it is necessary to retain the original features as long as a consistent mapping is created. In FastLGS and FMLGS, no compression is implemented and they have the first-tier performance.


\textbf{Limitation.} 
This method relies on accurate and consistent masks. Although the current segmentation models can easily provide object-level masks, generating masks for irregular common object parts remains a challenge. 
For some parts, even if they are successfully segmented in most views, they could still be ignored in some views where the parts are too far away from the camera, leading to inconsistency and degraded results (eg. some buttons). 
Therefore, in real-world applications, a single reconstruction may not be enough to obtain all consistent and accurate part-level information (depending on how detailed the scene is captured). 
For such scenarios, a close-up recapture and update should be necessary for demanding part-level interaction tasks. 

Meanwhile, although this language embedded gaussians ground detailed semantics for each target, no natural connection has been established within the fields. Future tasks will require the model to understand beyond simple target description, such as ``give me the largest bottle placed on the right of the book shelf''. 
Later works in this field may need to combine the vision language model to fill this gap.

\section{Conclusion}
In this paper, we present FMLGS, an approach that supports part-level open-vocabulary query within 3D Gaussian Splatting (3DGS). 
We propose an efficient pipeline for building and querying consistent object- and part-level semantics based on the Segment Anything Model 2 (SAM2). 
We also designed a semantic deviation strategy to solve the problem of language ambiguity among object parts. 
Once trained, FMLGS can query both objects and their describable parts using natural language. 
Comparisons with other state-of-the-art methods prove that our method can not only better locate specified part-level targets, but also achieve first-place performance concerning both \textbf{speed} and \textbf{accuracy}. 
Moreover,  we further designed an agent framework that integrates FMLGS as a virtual assistant that can interactively locate targets and respond to user demands within 3D scenes, 
showcasing the potential of our work to be further expanded and applied in the future.




\bibliography{IEEEtran}
\bibliographystyle{IEEEtran}

\newpage

\begin{IEEEbiography}[{\includegraphics[width=1in,height=1.25in,clip,keepaspectratio]{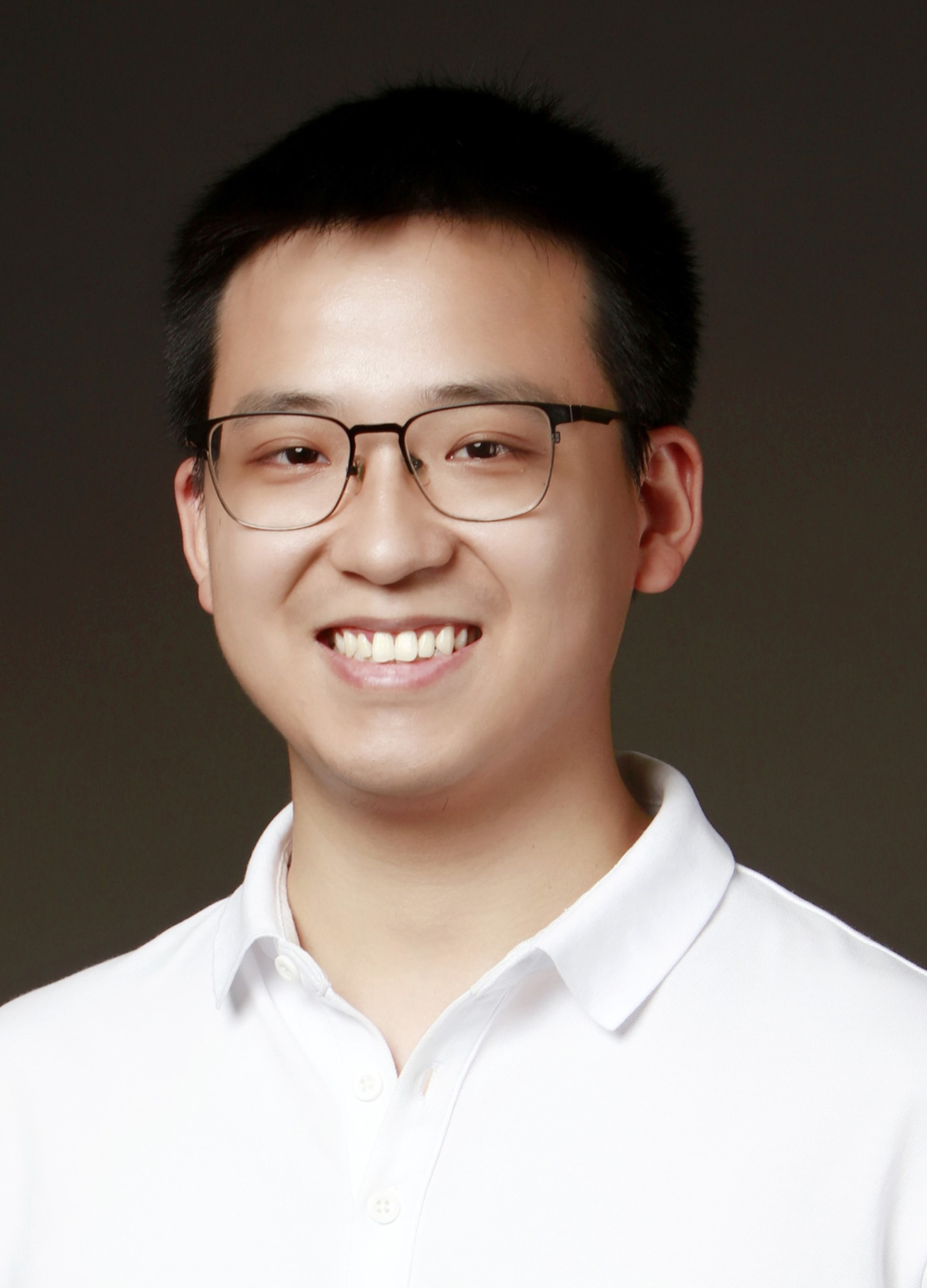}}]{Xin Tan} is currently the Research Professor (Zijiang Young Scholar) with School of Computer Science and Technology, East China Normal University, China. Before that, he was the Associate Research Professor at ECNU.
He received dual Ph.D. degrees in Computer Science from Shanghai Jiao Tong University and City University of Hong Kong in 2022. He received his B.Eng. degree in Automation from Chongqing University, China in 2017.
His research interests lie in computer vision and deep learning. He serves as a program committee member/reviewer for CVPR, ICCV, ECCV, AAAI, IJCAI, IEEE TPAMI, TIP and IJCV. He was selected for the Young Elite Scientists Sponsorship Program by CAST. He also serves as the associate editor for Pattern Recognition and Visual Computer.
\end{IEEEbiography}

\begin{IEEEbiography}[{\includegraphics[width=1in,height=1.25in,clip,keepaspectratio]{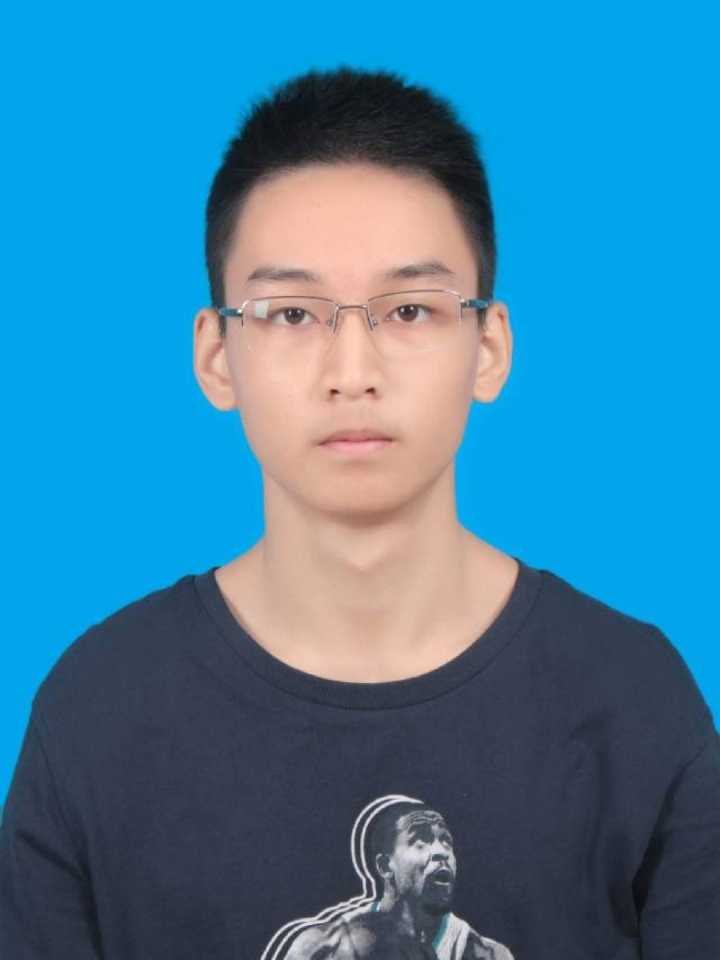}}]{Yuzhou Ji} is currently a forth-year undergraduate student at the School of Computer Science and Technology, East China Normal University, China. He is going to pursue his master degree at the Department of Computer Science and Engineering, Shanghai Jiao Tong University in 2025. His research interests cover 3D reconstruction and scene understanding.
\end{IEEEbiography}

\begin{IEEEbiography}[{\includegraphics[width=1in,height=1.25in,clip,keepaspectratio]{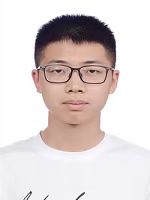}}]{He Zhu} is now a forth-year undergraduate student in the School of Computer Science and Technology, East China Normal University. He is going to pursue a master's degree in Electronic Engineering at Shanghai Jiao Tong University in 2025. His research interests cover scene reconstruction and neural rendering. 
\end{IEEEbiography}

\begin{IEEEbiography}[{\includegraphics[width=1in,height=1.25in,clip,keepaspectratio]{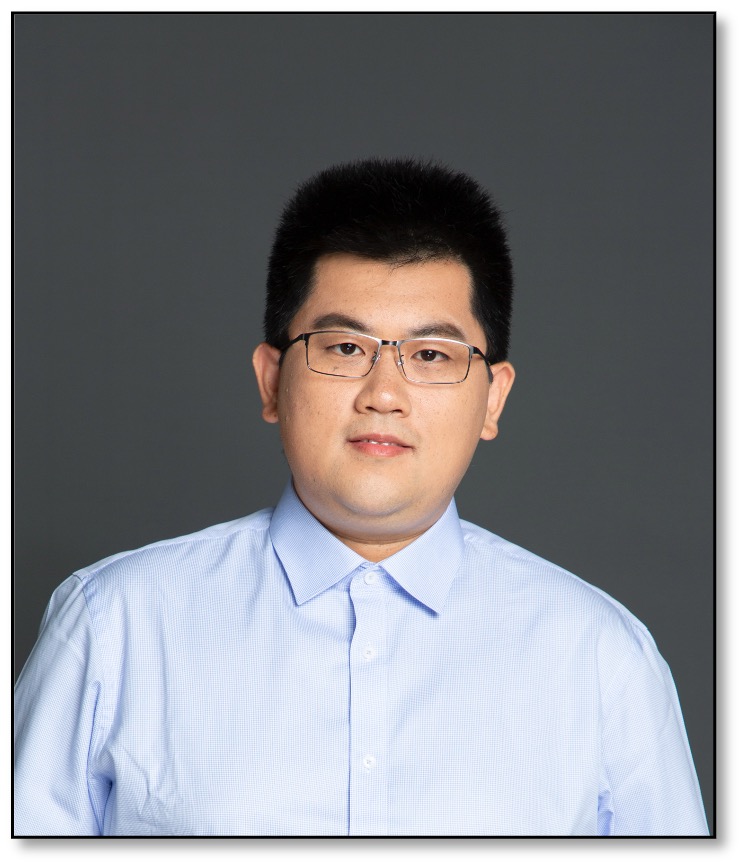}}]{Yuan Xie} received the PhD degree in
Pattern Recognition and Intelligent Systems from
the Institute of Automation, Chinese Academy of
Sciences (CAS), in 2013. He is currently a full
professor with the School of Computer Science and
Technology, East China Normal University, Shanghai,
China. His research interests include image
processing, computer vision, machine learning and
pattern recognition. He has published around 85 papers
in major international journals and conferences
including the IJCV, IEEE TPAMI, TIP, TNNLS,
TCYB, and NIPS, ICML, CVPR, ECCV, ICCV, etc.
He also has served as a reviewer for more than 15 journals and conferences.
Dr. Xie received the National Science Fund for Excellent Young Scholars 2022.
\end{IEEEbiography}

\vfill

\end{document}